\pdfoutput=1
\documentclass[11pt]{article}

\usepackage[preprint]{acl}

\usepackage{array}
\usepackage{multirow}
\usepackage{adjustbox}

\usepackage{booktabs} 
\usepackage{times}
\usepackage{latexsym}
\usepackage{amsmath}
\usepackage{natbib}
\usepackage{comment}
\usepackage[T1]{fontenc}
\usepackage[utf8]{inputenc}

\usepackage{microtype}

\usepackage{inconsolata}

\usepackage{graphicx}

\title{Conversational Query Reformulation with the Guidance of Retrieved Documents}

\author{Jeonghyun Park \and Hwanhee Lee\textsuperscript{$\dagger$} \\
    Department of Artificial Intelligence, Chung-Ang University, Seoul, Korea\\
    \texttt{\{tom0365, hwanheelee\}@cau.ac.kr}
}

\begin{document}
\maketitle
\footnotetext{\textsuperscript{$\dagger$}Corresponding author.}

\begin{abstract}
Conversational search seeks to retrieve relevant passages for the given questions in conversational question answering. Conversational Query Reformulation (CQR) improves conversational search by refining the original queries into de-contextualized forms to resolve the issues in the original queries, such as omissions and coreferences. Previous CQR methods focus on imitating human written queries which may not always yield meaningful search results for the retriever. In this paper, we introduce \textit{GuideCQR}, a framework that refines queries for CQR by leveraging key information from the initially retrieved documents. 
Specifically, \textit{GuideCQR} extracts keywords and generates expected answers from the retrieved documents, then unifies them with the queries after filtering to add useful information that enhances the search process.
Experimental results demonstrate that our proposed method achieves state-of-the-art performance across multiple datasets, outperforming previous CQR methods.
Additionally, we show that \textit{GuideCQR} can get additional performance gains in conversational search using various types of queries, even for queries written by humans.
\footnote{The code and pre-trained model will be publicly released upon the acceptance of the paper.}

\end{abstract}

\section{Introduction}

In a conversational question-answering task (ConvQA), conversational search aims to retrieve relevant passages that provide the necessary information to answer the current query. This process occurs within the framework of a multi-turn conversation, where each query builds upon the previous interactions.
The questions in ConvQA often involve challenges like omissions and coreferences, which make it difficult to achieve the desired search results using the original query.
Previous research focuses on transforming queries into stand-alone forms to understand their intent better, making them more independent and robust~\cite{mo2023convgqr}. This process, known as Conversational Query Reformulation (CQR), helps clarify the original queries and enhances query understanding.

\begin{figure}[t]
\centering
\includegraphics[width=0.98\columnwidth]{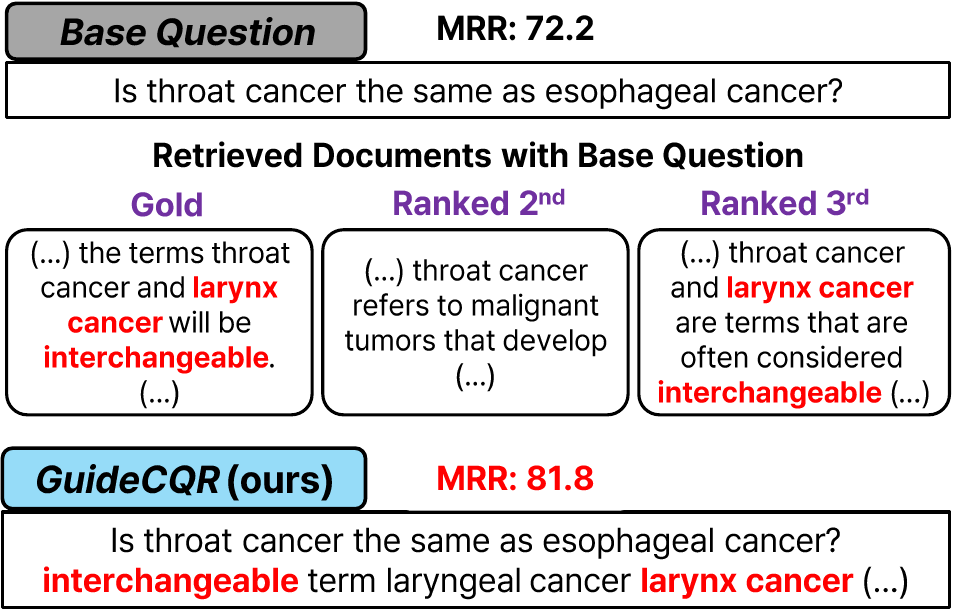}
\caption{Example queries for ConvQA where the reformulated query achieves a higher MRR score by effectively extracting clues from the initially retrieved documents, compared to the base query.}
\vspace{-5mm}
\label{fig:motivation}
\end{figure}

With the advent of Large Language Models (LLMs), their application in CQR methods has become increasingly widespread.
A recent study~\cite{jagerman2023query} involves instructing an LLM to generate relevant passages related to the query. This method makes the query more reasonable and human-understandable by expanding the sentence. LLM4CS~\cite{mao2023large} also involves using LLMs to rewrite the query through different prompting methods and aggregation techniques.
However, although these approaches aim to create human-friendly and easily comprehensible queries, they may not always yield the effective results that retrievers desire. 
For instance, as shown at the bottom of Figure~\ref{fig:motivation}, although the reformulated query may seem less fluent due to the additional keywords, it achieves a higher retriever score than a base query because of its increased similarity to the ground-truth document.
This example highlights the necessity of prioritizing retriever-optimized queries over human-friendly ones to enhance CQR methods. To achieve this, we observe that the initially retrieved documents from a retrieval process with a base query set can help the generation of retrieval-optimized queries.
Specifically, we find that although some of the documents cannot provide clues for answering the questions, most of the retrieved documents contain words or signals that are highly influential in subsequent retrieval processes.
For instance, in the example illustrated in Figure ~\ref{fig:motivation}, the addition of the terms \textit{'interchangeable'}, an important keyword found in the gold passage within the guided documents, enhances the performance that the original query alone cannot achieve.
In this way, signals in the initially retrieved documents obtained through a base query can guide the search for the gold passage.

In this paper, we propose \textit{GuideCQR}, an effective CQR framework designed to generate retrieval-friendly conversational queries by leveraging the guidance of initially retrieved documents from the base query.
\textit{GuideCQR} first obtains the initially retrieved documents through a retrieval process using the baseline query reformulated by LLM. 
We then perform a re-ranking process to refine the order of the retrieved documents based on their similarity to the query. Through this re-ranking process, we carefully select a small number of documents from the initially retrieved ones to extract signals that are more likely to contain the gold passage.
Next, we extract the keywords and generate the expected answer from the re-ranked guided documents to obtain components for creating a retriever-friendly query. 
We then independently filter both the keywords and the expected answers based on their similarity to the original query and previous utterances to eliminate redundant information. Finally, we construct the final query set by concatenating the filtered keywords and expected answers with the baseline query.

Experimental results show that \textit{GuideCQR} achieves state-of-the-art across several CQR datasets compared to the baseline systems.
We also demonstrate the robustness of the \textit{GuideCQR} framework in enhancing the retrievability of queries, making them more retriever-friendly. 
Furthermore, we validate the effectiveness of our method by analyzing the overlap between the augmented keywords and the relevant documents for the given question.
Additionally, we show that the proposed method achieves notable improvements across various query sets, underscoring its adaptability.

\begin{figure*}[!h]
\centering
\includegraphics[width=1.98\columnwidth]{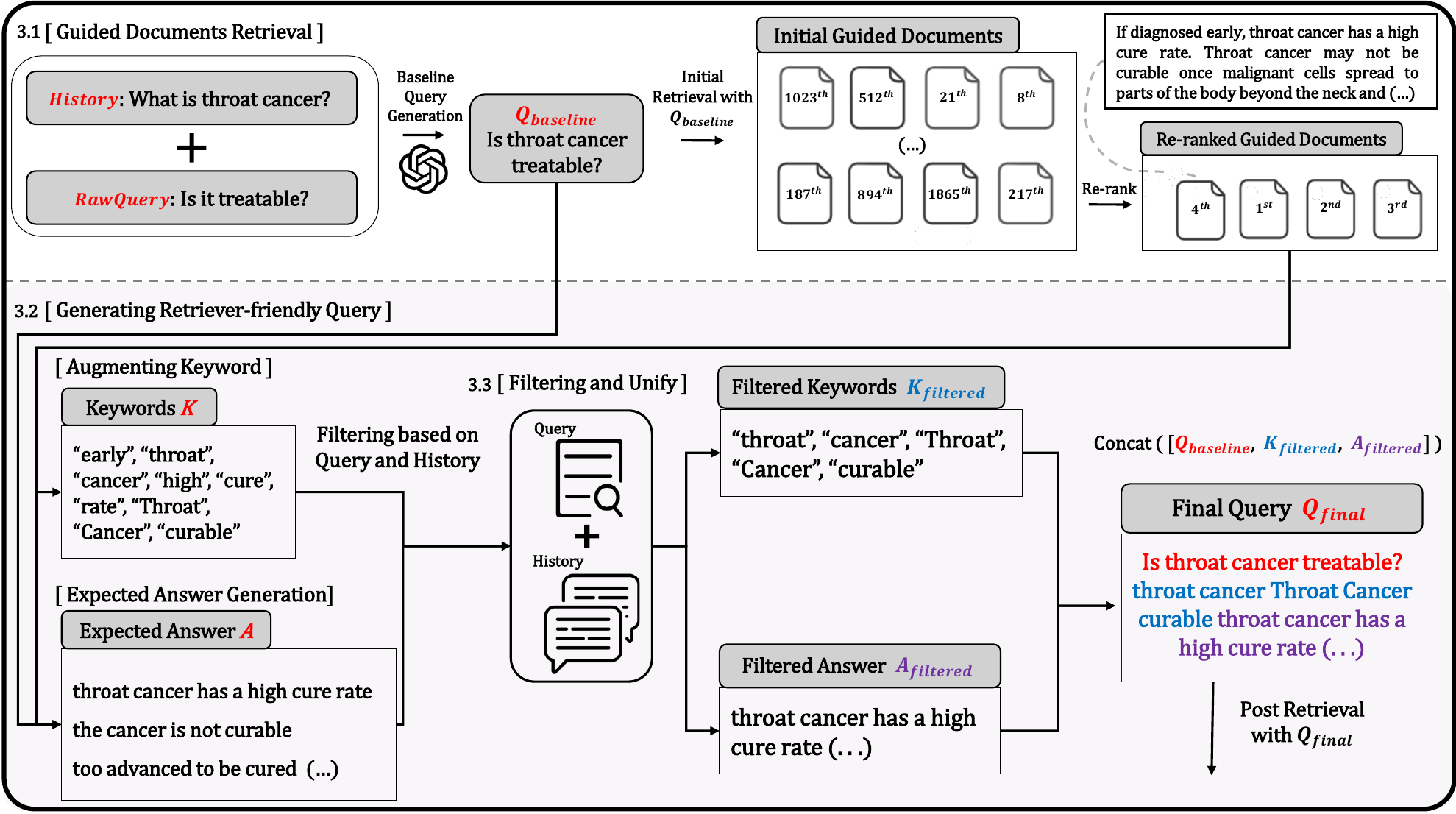}
\caption{Overall framework of \textit{GuideCQR}: For easier understanding, we only visualize top-1 ranked document and present keywords augmented from the top-1 document and 3 answer pairs from the top-3 documents.}
\vspace{-2mm}
\label{fig:framework}
\end{figure*}

\section{Related Works}

\subsection{Conversational Search}
The goal of conversational search ~\cite{gao2023neural} is to retrieve passages containing the necessary information to answer the current query within a multi-turn conversation. To get the desired answers from the given conversational dialogue, understanding the meaning of the query is important in this field ~\cite{mo2023learning}. However, queries in ConvQA have several problems. For instance, queries may include pronouns such as "he," "she," or "it," necessitating the identification of the referred entity, and there is also the challenge of omission, where essential information is not included~\cite{mao2023large, wang2023query2doc}. 

To address such challenges in conversational search, two main methods that enable conversational search are Conversational Dense Retrieval (CDR) and CQR. 
CDR focuses on enhancing the representation of the current query by incorporating historical context achieved through training dense retrievers~\cite{qu2020open}. On the other hand, CQR transforms the conversational search into a traditional ad-hoc search by converting the entire search session into a single standalone query~\cite{elgohary2019can}.

In our work, we focus on utilizing the dialogue history to contextualize query rewrites. Additionally, we employ the multi-turn dialogue as a criterion for filtering keywords and answers, which are essential components of \textit{GuideCQR}.

\subsection{Conversational Query Reformulation}
CQR methods aim to transform queries into de-contextualized forms, enabling them to be understood based solely on their content and ensuring they convey the intended meaning effectively. These methods are designed to enhance conversational search performance by refining and expanding user queries within a conversational context.

While previous approaches, such as human-rewritten queries~\cite{sheng2020conversational, voskarides2020query, yu2020few}, have attempted to improve query clarity, they often fall short in optimizing retrieval performance. Although human rewrites may enhance readability for users, they do not always align with the needs of retrieval systems.

To address these challenges, recent studies have introduced methods like CONQRR~\cite{wu2022conqrr}, which directly optimizes query rewriting for retrieval through Self-Critical Sequence Training~\cite{rennie2017self}, and ConvGQR\cite{mo2023convgqr}, which improves retrieval performance by combining query rewriting with query expansion. Additionally, IterCQR~\cite{jang-etal-2024-itercqr} iteratively trains the CQR model by using information retrieval (IR) signals as rewards. The integration of LLMs in CQR has also shown promise in recent work~\cite{mao2023large}.

To the best of our knowledge, no prior study has directly utilized the signals within retrieved documents to apply CQR. Our proposed method, \textit{GuideCQR}, focuses on transforming queries by leveraging information from the retrieved documents alongside the baseline query during the retrieval process.

\section{GuideCQR} 
We propose \textit{GuideCQR}, a novel framework designed to reformulate conversational queries by utilizing guidance from initially retrieved documents. Figure~\ref{fig:framework} provides an overview of the proposed query reformulation process.
\textit{GuideCQR} consists of three stages: We first retrieve an initial set of guided documents using a query reformulated by LLM.
This step retrieves documents that are likely to contain gold passages to guide the query reformulation process (Sec~\ref{sec:1_guided}). 
Next, we extract keywords and generate expected answers from the guided documents, creating components that contribute to making the query more retriever-friendly (Sec~\ref{sec:3_generate}). 
Finally, we apply a filtering process that evaluates both keywords and answers based on their similarity scores relative to the baseline query and dialogue history. 
We then unify and concatenate these components with the baseline query to construct the final query for post retrieval (Sec~\ref{sec:4_filter}).

\subsection{Guided Documents Retrieval}
\label{sec:1_guided}

\subsubsection{Initial Documents Retrieval} 
In the process of developing a retriever-friendly query, initially retrieved documents can play a crucial role as guiding resources by providing foundational insights for the CQR process. For example, in the contents of retrieved documents, these signals can include critical keywords or contextual signs that are necessary to the search the gold passages such as "cure," "rate," and "curable" from the document shown in Figure ~\ref{fig:framework}. 

Inspired by these points, \textit{GuideCQR} starts from getting initially retrieved documents to gain meaningful signals to the retriever. We obtain these initial documents by retrieving the documents using the baseline query set generated by LLM. We denote baseline query as $Q_{baseline}$, 
\begin{equation} 
\footnotesize Q_{baseline} = Rewrite_{LLM} (History, RawQuery),
\end{equation} 
where $Rewrite_{LLM}$ represents an operation that resolves omissions or coreferences using OpenAI \textit{gpt3.5-turbo-16k} and $RawQuery$ denotes the raw question in the dataset, without any reformulation applied. $History$ denotes the dialogue history of $RawQuery$, especially consisting of previous dialogues' queries, rewritten queries and responses by human annotators. Based on both the $RawQuery$ and $History$, we generate the $Q_{baseline}$. Using this $Q_{baseline}$, we obtain the initial documents composed of 2,000 documents for each question. They are crucial for creating our final query set and strong foundation for identifying meaningful signals and guiding the retriever towards the most relevant passages.

\subsubsection{Re-ranking Documents}
\label{sec:rerank}
To further improve the quality of the guided set of documents, we implement a re-ranking process to generate initially retrieved documents. By reordering the initial documents using a different retrieval model, we aim to capture better documents by reducing biases that may arise from relying on a single retriever. 
Specifically, we employ Sentence-Transformer~\cite{reimers2019sentence} to re-rank the top 2,000 documents for each question, selecting the final guided set by choosing the top 10 documents based on their similarity scores to the query.

\subsection{Generating Retriever-friendly Query}
\label{sec:3_generate}
Since many retrieved documents may contain influential words or signals that significantly impact subsequent retrieval stages, identifying key elements from these documents is crucial for constructing more effective queries.
Based on the guided documents obtained from the previous step, we generate retriever-friendly queries by incorporating extra information to the query from two approaches: \textit{Augmenting Keywords}, \textit{Expected Answer generation}

\subsubsection{Augmenting Keywords} 
We find that keywords from the initial documents play a critical role in forming retriever-friendly queries, as they capture the most relevant and significant terms within the document. For example, keywords such as "early", "throat", "cancer", "high", and "cure" can be beneficial if they are augmented to the search query as shown in Figure~\ref{fig:framework}.% we utilize the guided documents.

%We extract these keywords from the initial documents by leveraging BERT~\cite{devlin-etal-2019-bert} through the keyword extraction method as in KeyBERT~\cite{grootendorst2020keybert}. 
%We leverage BERT~\cite{devlin-etal-2019-bert} to extract keywords from the initial documents using a keyword extraction method similar to KeyBERT~\cite{grootendorst2020keybert} as follows.

We leverage KeyBERT~\cite{grootendorst2020keybert} to extract keywords from the re-ranked documents.
The process begins by using BERT~\cite{devlin-etal-2019-bert} to compute embeddings of documents, which create a representation for the entire document. Embeddings for N-gram words or phrases within the document are then extracted. By calculating the cosine similarity between these words/phrases and the document representation, the method identifies which words/phrases are most similar to the document.

%We use BERT to compute embeddings, creating a representation for the entire document. Subsequently, we extract embeddings for N-gram words/phrases in the document. By computing cosine similarity between words and document, we determine which words/phrases are most similar to the document. We choose the keywords with the highest similarity. 

In line with this principle, we enhance the base queries by augmenting keywords through two hyperparameters.
%The first aspect concerns determining the number of documents to extract keywords from, which defines the level of guidance we aim to achieve. For each document, we extract keywords of a specified span length. The second aspect focuses on the span length of the keywords, which refers to the number of tokens, or the number of keywords, selected per document. For example, when augmenting top-2 span-3 keywords, we extract three keywords from each of the top two documents, resulting in a total of six keywords.
The first one involves determining the number of documents to extract keywords from, which establishes the level of guidance we intend to provide. For each guided document, we extract keywords of a specified span length. The second aspect pertains to the span length, which indicates the number of tokens or keywords selected per document. For instance, when augmenting with top-2 span-3 keywords, we extract three keywords from each of the top two documents, yielding a total of six keywords.

Consequently, we augment a total keyword list $K$ composed of $nm$ keywords from the top-$n$ documents and span length $m$: 
\begin{equation} K = [k_{11}, k_{12}, k_{13}, ... , k_{1m}, k_{21}, ..., k_{nm}], 
\end{equation}
where $k$ denotes unit keyword, $n$ is the number of documents and $m$ is keyword span length. 

\subsubsection{Expected Answer Generation} 
Guided documents often include gold answers to the query, serving as a valuable resource for efficiently reformulating the query. Based on this idea, we use guided documents as context to generate expected answers to enhance the query.
%Similar to keywords, we also observe that the expected answers to queries obtained through guided documents can further assist in finding gold passages.
As illustrated in Figure~\ref{fig:framework}, although it's not obtained from the gold passage, expected answer is a concise and informative response that can potentially address the user's query since they can provide direct insights into the user's intent, thereby improving the relevance and accuracy of search results. Specifically, we generate these expected answers as follows:
%Similar to keywords, expected answers can also serve as valuable signals to better understand and refine the query.

%Similar to keywords, we also find that expected answers for the given query such as "throat cancer has a high cure rate", as illustrated in Figure~\ref{fig:framework} can also serve as valuable signals to the query. 

%We find that when generating the expected answer, rather than instructing language models to generate potential answers to the query, guided documents may contain gold answers to the query and can play a crucial role as a pool for finding the answers to reformulate query.
\begin{equation} 
A = [a_1, a_2, a_3, ..., a_n], 
\end{equation} 
where $A$ represents answer list and $a_{n}$ denotes a unit answer extracted from a single document. 
To generate these answers, both the query and relevant context are necessary.
We use the query as the baseline query $Q_{baseline}$ and the context as the guided documents. We generate a single expected answer for each document. Consequently, this process produces k distinct answers, with one derived from each of the top-k documents.

\subsection{Filtering and Unify}
\label{sec:4_filter}
We observe that redundant elements, such as the keyword \textit{'rate'} as shown in Figure~\ref{fig:framework}, can emerge from both the augmented keywords and the generated answers.
%We explain this is because \textit{GuideCQR} might augment the keywords or answers generated from irrelevant documents. 
We find that this is because \textit{GuideCQR} might augment keywords or answers derived from irrelevant documents. Hence, irrelevant signals from irrelevant documents can have a negative impact when creating retriever-friendly queries. 

To address this, we introduce an additional filtering stage to more effectively remove irrelevant keywords and answers from the reformulated query. We guide this filtering process through the metric \textit{FilterScore}, which leverages both $QueryScore$ and $HistoryScore$, calculated using cosine similarity, as follows:

\begin{equation}
cosSim(x,y) = \frac{x \cdot y}{\|x\| \|y\|},
\end{equation}

\noindent where \textit{cosSim} denotes cosine similarity ranging from 0 to 1. Based on $cosSim$, we firstly define $QueryScore$ as follows:
\begin{equation}
\textit{QueryScore} = 10 \cdot cosSim(\textit{query}, \textit{item}),
\end{equation}

\noindent where \textit{query} represents the embedding of the current turn's query, and \textit{item} refers to the embedding of either a keyword or an answer sentence. So $QueryScore$ is cosine similarity between query and item ranging from 0 to 10. And we define $HistoryScore$:

{\footnotesize
\begin{equation} 
\textit{HistoryScore} = 10 \cdot \max \left( cosSim(\textit{history}[i], \textit{item}) \right),
\end{equation}}

\noindent where \textit{history} is the history query list of current utterances. So \textit{HistoryScore} is the maximum cosine similarity value between the dialogue history query element in the history query list and current item. Finally, we define the \textit{FilterScore} as the average of the \textit{QueryScore} and the \textit{HistoryScore}, ranging from 1 to 10:

\begin{equation} 
\textit{FilterScore} = \frac{\textit{QueryScore} + \textit{HistoryScore}}{2}.
\end{equation}

History queries play a crucial role in understanding the current query, as in a dialogue, comprehending the question is more effective when based on the preceding conversation. 
Thus, we account for history queries through the $HistoryScore$, which allows us to traverse the entire dialogue and capture the global context from past to present. 
Using this $FilterScore$, we can eliminate signals that are irrelevant to the current dialogue. 
Since keywords and answers may vary in retriever-friendliness across different datasets, we treat them as distinct units and apply different filtering scores rather than using the same score for both.

Using the $FilterScore$, we filter out keywords and answers with a score below the specified threshold. Finally, we unify the remaining keywords and answers and integrate them into the $Q_{baseline}$ to construct the final reformulated query as follows:

{\small
\begin{equation} 
Q_{final} = Concat([Q_{baseline}, K_{filtered}, A_{filtered}]), 
\end{equation}}
where $K_{filtered}$ and $A_{filtered}$ are remaining keywords, answers.

\begin{table*}[h]
    \centering
    \small
    \begin{tabular}{l|ccc|ccc|ccc|c}
        \toprule
        & \multicolumn{3}{c|}{CAsT-19} & \multicolumn{3}{c|}{CAsT-20} & \multicolumn{3}{c|}{QReCC} & \\
        \cmidrule(lr){2-4} \cmidrule(lr){5-7} \cmidrule(lr){8-10} 
        \textbf{Methods} & MRR & NDCG@3 & R@10 & MRR & NDCG@3 & R@10 & MRR & NDCG@3 & R@10 & \textbf{Avg} \\
        \midrule
        $RawQuery$ & 41.2 & 24.3 & 5.8 & 23.2 & 15.4 & 5.5 & 10.2 & 9.3 & 15.7 & 16.7\\
        Transformer++ & 69.6 & 44.1 & - & 29.6 & 18.5 & - & - & - & - & - \\
        CQE-Sparse & 67.1 & 42.3 & - & 42.3 & 27.1 & - & 32.0 & 30.1 & 51.3 & - \\
        QuReTeC & 68.9 & 43.0 & - & 43.0 & 28.7 & - & 35.0 & 32.6 & 55.0 & - \\
        T5QR & 70.1 & 41.7 & - & 42.3 & 29.9 & - & 34.5 & 31.8 & 53.1 & - \\
        ConvGQR & 70.8 & 43.4 & - & 46.5 & 33.1 & - & 42.0 & 39.1 & 63.5 & - \\
        CONVINV & 74.2 & 44.9 & - & 47.6 & 34.4 & - & - & - & - & - \\
        CHIQ & 73.3 & 50.5 & \underline{12.9} & 54.0 & 38.0 & \underline{19.3} & \underline{47.0} & 44.2 & \textbf{70.8} & 45.5 \\
        IterCQR	& - & - & - & - & - & - & 42.9 & 40.2 & 65.5 & - \\
        \midrule
        $Q_{baseline}$ & 72.2 & 45.2 & 12.2 & 53.4 & 37.2 & 18.2 & 43.8 & \underline{51.9} & 66.4 & 44.5 \\
        LLM4CS & \underline{77.6} & \underline{51.5} & 10.9 & \textbf{61.5} & \textbf{45.5} & 17.0 & 44.8 & 42.1 & 66.4 & \underline{46.3} \\
        \textbf{GuideCQR (ours)} & \textbf{81.8} & \textbf{53.7} & \textbf{13.9} & \underline{59.0} & \underline{42.7} & \textbf{21.1} & \textbf{47.2} & \textbf{54.4} & \underline{67.0} & \textbf{48.9} \\
        \midrule
        Human Rewrite & 74.0 & 46.1 & 12.9 & 59.1 & 42.2 & 21.0 & 43.1 & 51.6 & 58.6 & 45.4 \\
        \bottomrule
    \end{tabular}
    \caption{Performance comparison for conversational search on various CQR methods on CAsT-19, CAsT-20 and QReCC dataset. We present MRR, NDCG@3, R@10, and the average of all scores for each method. The best results are in bold, and the second best are underlined. The dashes ('-') indicate performance values that could not be measured due to differences in evaluation metrics or the unavailability of results from closed-source systems.}
    \label{tab:1}
    \vspace{-1mm}
\end{table*}

\section{Experiments}
\subsection{Datasets and Metrics}

We utilize three CQR benchmark datasets TREC CAsT-19~\citep{dalton2020cast}, TREC CAsT-20~\citep{dalton2021cast} and QReCC dataset~\cite{anantha-etal-2021-open} for our work. 
%CAsT datasets are curated by human experts from the TREC Conversational Assistance Track (CAsT). 
CAsT-19 and CAsT-20 consist of 50 and 25 conversations, respectively. Both CAsT datasets share the same document collection and provide passage-level relevance judgments, as well as human rewrites for each turn. Unlike CAsT-19, CAsT-20 is more realistic and complex because its queries are based on information needs drawn from commercial search logs, and they can reference prior system responses.
CAsT series datasets assign a relevance score ranging from 1 to 4 to each passage, indicating the degree of relevance to the query, with documents scoring a 4 being considered gold passages.
For the QReCC dataset, each query is paired with a single gold passage different from CAsT dataset.
QReCC dataset includes a training set and a test set, and we sample 2K conversations from the training set to create a development set, following the previous work~\cite{kim2022saving}.

Following the prior CQR research~\citep{mo2023convgqr, mao2023large, jang-etal-2024-itercqr}, we use three widely used evaluation metrics for CQR to compare the performance: Mean Reciprocal Rank (MRR), Normalized Discounted Cumulative Gain at three documents (NDCG@3), Recall@10. We utilize \textit{pytrec\_eval}~\citep{van2018pytrec_eval} tool to compute the score. 

\subsection{Implementation Details} 
To generate a baseline query $Q_{baseline}$ for CAsT-19 and CAsT-20, we utilize OpenAI \textit{gpt3.5-turbo-16k} combined with the Maxprob approach as proposed in LLM4CS. 
We simply generate this query by instructing LLM to rewrite the raw query based on dialog history and sampling the highest generation probability with LLMs, resolving only omissions and coreferences. 
For the QReCC dataset, rather than generating $Q_{baseline}$ ourselves, we adopt the final query output generated by InfoCQR~\cite{ye-etal-2023-enhancing} as $Q_{baseline}$. This approach leverages prompts from \textit{gpt3.5-turbo} to enhance query generation. 

Following previous studies~\cite{mao2023large, jang-etal-2024-itercqr}, we use ANCE~\cite{xiong2020approximate} pre-trained on the MSMARCO~\cite{Campos2016MSMA} as our retriever. For further implementation details, please refer to Appendix~\ref{app:details}.
                
\subsection{Baselines}
We compare \textit{GuideCQR} with the following CQR methods: (1) \textbf{Transformer++}~\citep{svitlana2021ain}: A GPT-2 based CQR model fine-tuned on CANARD~\cite{elgohary2019can} dataset. (2) \textbf{CQE-Sparse}~\citep{sheng2021bin}: A weakly-supervised method to select important tokens only from the context via contextualized query embeddings. (3) \textbf{QuReTeC}~\citep{nikos2020in}: A weakly-supervised method to train a sequence tagger to decide whether each term contained in a historical context should be added to the current query. (4) \textbf{T5QR}~\citep{sheng2020conversational}: A conversational query rewriter based on T5, trained using human-generated rewrites. (5) \textbf{ConvGQR}~\citep{mo2023convgqr}: A query reformulation framework that combines query rewriting with generative query expansion. (6) \textbf{CONVINV}~\citep{cheng-etal-2024-interpreting}: Framework that transforms conversational session embeddings into interpretable text using Vec2Text. (7) \textbf{CHIQ}~\cite{mo2024chiq}: A two-step method that leverages the capabilities of LLMs to resolve ambiguities in the conversation history before query rewriting. (8) \textbf{IterCQR}~\cite{jang-etal-2024-itercqr}: CQR framework through iterative refinement based on the similarity between the passage and the query. (9) \textbf{LLM4CS}: Query rewriting based on LLM and various prompting methods. 
%We build \textit{GuideCQR} upon this prompted QR method using (REW + Maxprob) and evaluate using self-consistency. 

\subsection{Performance Comparison}
\paragraph{Main Results}
As shown in Table~\ref{tab:1}, \textit{GuideCQR} significantly enhances performance metrics compared to the $Q_{baseline}$ across all datasets. \textit{GuideCQR} achieves state-of-the-art performance in terms of average score. In addition, \textit{GuideCQR} achieves either the best or second-best performance compared to all baselines and demonstrates its robustness. Specifically, \textit{GuideCQR} outperforms \textbf{LLM4CS} by 5.4\% in MRR on the CAsT-19 dataset and by 29.2\% in NDCG@3 on the QReCC dataset. For CAsT-20, still its performance is almost the second-best. These results highlight the robustness and effectiveness of \textit{GuideCQR}.

We attribute the lower score for CAsT-20 compared to CAsT-19 to the increased complexity of the topics in CAsT-20.
Specifically, in the retrieval process, we set the parameter \textit{rel threshold} which represents the minimum query relevance score for a document to be considered relevant to the query. CAsT-19 uses this threshold as 1 and CAsT-20 uses 2, so the minimum criteria for relevance is higher in CAsT-20. As a result, the number of relevant documents in CAsT-20 is significantly lower than in CAsT-19. 
Furthermore, the query relevance score file for CAsT-20 contains relatively few documents with high relevance scores; in most cases, the score is 0. This makes CAsT-20 a more challenging dataset compared to CAsT-19 for applying \textit{GuideCQR}. In other words, the signals derived from irrelevant guiding documents may not perform effectively to \textit{GuideCQR} in CAsT-20.

\paragraph{Ablation Study}
To evaluate the effectiveness of individual components involved in creating retriever-friendly queries in our proposed CQR framework, we conduct ablation study. As demonstrated in Table~\ref{tab:ablation}, removing any part of \textit{GuideCQR} leads to a decrease in performance, indicating that each component plays an important role in \textit{GuideCQR}.

\begin{table}[h]
    \centering
    \resizebox{\columnwidth}{!}{ 
    \begin{tabular}{l|cc|cc}
        \toprule
        & \multicolumn{2}{c|}{CAsT-19} & \multicolumn{2}{c}{CAsT-20} \\
        \cmidrule(lr){2-3} \cmidrule(lr){4-5}
        & MRR & NDCG@3 & MRR & NDCG@3 \\
        \midrule
        \textbf {GuideCQR (ours)} & 81.8 & 53.7 & 59.0 & 42.7 \\
        \ \ \ - w/o keywords & 72.0 & 47.3 & 54.7 & 38.6 \\
        \ \ \ - w/o answers & 81.0 & 52.2 & 57.9 & 41.6 \\
        \ \ \ - w/o filtering & 79.7 & 52.2 & 58.4 & 43.2 \\
        \bottomrule
    \end{tabular}
    }
    \caption{Ablation study on each component of \textit{GuideCQR}.}
    \label{tab:ablation}
    \vspace{-2mm}
\end{table}
%Caption 수정 

\subsection{Analysis}
CAsT dataset includes multiple relevant passages, each with a relevance score. We focus on CAsT due to these unique label features.

\paragraph{Performance among Top-k Documents for Keyword and Answer}
As shown in Table~\ref{tab:3}, we verify the impact of varying the number of documents used in augmenting keywords and generating expected answers. Our findings indicate that incorporating a larger number of documents generally provides additional information, which can improve performance. However, using too many documents may cause a decline in overall performance similar to the amount of initial guided documents.

\begin{table}[!ht]
    \centering
    \resizebox{\columnwidth}{!}{ 
    \begin{tabular}{l|ccc|ccc}
        \toprule
        & \multicolumn{3}{c|}{Keyword} & \multicolumn{3}{c}{Answer} \\
        \cmidrule(lr){2-4} \cmidrule(lr){5-7}
        & MRR & NDCG@3 & Avg & MRR & NDCG@3 & Avg \\
        \midrule
        $Q_{baseline}$ & 72.2 & 45.2 & 58.7 & 72.2 & 45.2 & 58.7 \\
        top-2 & 78.3 & 50.1 & 64.2 & 76.7 & 48.3 & 62.5 \\
        top-3 & 80.4 & 52.1 & 66.2 & 77.5 & 49.0 & 63.2 \\
        top-4 & 81.0 & 52.2 & 66.6 & 76.9 & 50.0 & 63.4 \\
        top-5 & 80.2 & 52.3 & 66.2 & 77.5 & 49.6 & 63.5 \\
        \bottomrule
    \end{tabular}
    }
    \caption{Performance on the CAsT-19 dataset with the different numbers of documents for extracting keywords. We augment span 15 keywords from each document.}
    \label{tab:3}
    \vspace{-0mm}
\end{table}

\paragraph{Performance among Amount of Initial Guided documents}

\begin{table}[h]
    \centering
    \resizebox{\columnwidth}{!}{ 
    \begin{tabular}{c|ccc}
        \toprule
        \# of initial guided documents & MRR & NDCG@3 & Avg \\
        \midrule
        10  & 75.7 & 50.5 & 63.1 \\
        100 & 78.2 & 51.4 & 64.8 \\
        1,000 & 79.5 & 52.1 & 65.8 \\
        2,000 & 81.0 & 52.2 & 66.6 \\
        \bottomrule
    \end{tabular}
    }
    \caption{Evaluation on the CAsT-19 dataset with varying amount of guided documents.}
    \label{tab:7}
    \vspace{-2mm}
\end{table}

We evaluate the impact and robustness of each step in the \textit{GuideCQR} setup. 
Initially, we adjust the number of guided documents to observe the proper quantity and present the results in Table~\ref{tab:7}. Our findings demonstrate that increasing the number of guided documents consistently enhances performance. However, retrieving an excessive number of documents leads to longer inference times. Hence, considering the computation cost, we decide to use 2000 documents which is the point on the dev set where there is no additional performance improvement in at least one metric. We also use the same criteria for keyword span length.

\begin{table*}[h!]
\centering
\small
\begin{adjustbox}{max width=\textwidth}
\begin{tabular}{lccc|ccc}
\toprule
\multirow{2}{*}{} & \multicolumn{3}{c|}{$Q_{baseline}$} & \multicolumn{3}{c}{GuideCQR} \\
\midrule
 & MRR & NDCG@3 & R@10 & MRR & NDCG@3 & R@10 \\
\midrule
RawQuery              & 41.2 & 24.3 & 5.8  & 47.0\textsubscript{\textcolor{red}{(+5.8)}} & 28.8\textsubscript{\textcolor{red}{(+4.5)}} & 6.6\textsubscript{\textcolor{red}{(+0.8)}} \\
Human Rewrite         & 74.0 & 46.1 & 12.9 & 82.8\textsubscript{\textcolor{red}{(+8.8)}} & 53.0\textsubscript{\textcolor{red}{(+6.9)}} & 14.4\textsubscript{\textcolor{red}{(+1.5)}} \\
GPT3.5-turbo-16k      & 72.2 & 45.2 & 12.2 & 81.8\textsubscript{\textcolor{red}{(+9.6)}} & 53.7\textsubscript{\textcolor{red}{(+8.5)}} & 13.9\textsubscript{\textcolor{red}{(+1.7)}} \\
Qwen2.5-7B-Instruct   & 62.9 & 39.3 & 10.2 & 65.0\textsubscript{\textcolor{red}{(+2.1)}} & 42.3\textsubscript{\textcolor{red}{(+3.0)}} & 11.7\textsubscript{\textcolor{red}{(+1.5)}} \\
Mistral-7B-Instruct-v0.3 & 69.1 & 42.8 & 11.6 & 69.8\textsubscript{\textcolor{red}{(+0.7)}} & 46.0\textsubscript{\textcolor{red}{(+3.2)}} & 12.0\textsubscript{\textcolor{red}{(+0.4)}} \\
EXAONE-3.5-7.8B-Instruct & 66.4 & 41.5 & 11.1 & 72.6\textsubscript{\textcolor{red}{(+6.2)}} & 46.7\textsubscript{\textcolor{red}{(+5.2)}} & 12.5\textsubscript{\textcolor{red}{(+1.4)}} \\
\bottomrule
\end{tabular}
\end{adjustbox}
\caption{Performance comparison of applying \textit{GuideCQR} on $RawQuery$, Human rewritten query and $Q_{baseline}$ with different generator models. Metrics include MRR, NDCG@3, and Recall@10, with performance improvements highlighted in red.}
\label{tab:adaptability}
\vspace{-4mm}
\end{table*} 

\paragraph{Query Relevance Analysis}

\newcommand{\NumFour}{K_{rel}} 
\newcommand{\TotalKeywords}{K_{\text{total}}} 

To demonstrate the effectiveness of the keyword augmentation step, we evaluate the precision score using the following formula:
\begin{equation} 
Precision = \frac{\NumFour}{\TotalKeywords}, 
\end{equation}
where $K_{rel}$ represents the number of unique matched keywords found in the guided documents that have query relevance score of $rel$ and $K_{total}$ represents the total number of unique augmented keywords. Note that the document that has a relevance score of 1 or is regarded as the most relevant document with user queries in CAsT datasets. 
Matching keywords in these documents can confirm that augmenting keywords plays a crucial role in generating retriever-friendly queries. As shown in Table~\ref{tab:6}, the precision score for relevant passages is higher than that for irrelevant passages, indicating that augmenting keywords play a crucial role in retrieving relevant documents.

\begin{table}[h!]
\centering
\small
\begin{tabular}{l l c}
\toprule
\textbf{} & \textbf{} & \textbf{Precision} \\
\midrule
Irrelevant & Passage (0) & 57.21 \\
\midrule
\multirow{4}{*}{Relevant} & Passage (1) & 67.39 \\
 & Passage (2) & 71.36 \\
 & Passage (3) & 72.26 \\
 & Passage (4) & 70.45 \\
\bottomrule
\end{tabular}
\caption{Precision for irrelevant and relevant passages of CAsT-19 dataset.}
\label{tab:6}
\vspace{-4mm}
\end{table}

\begin{figure}[!h]
\centering
\includegraphics[scale=0.48]{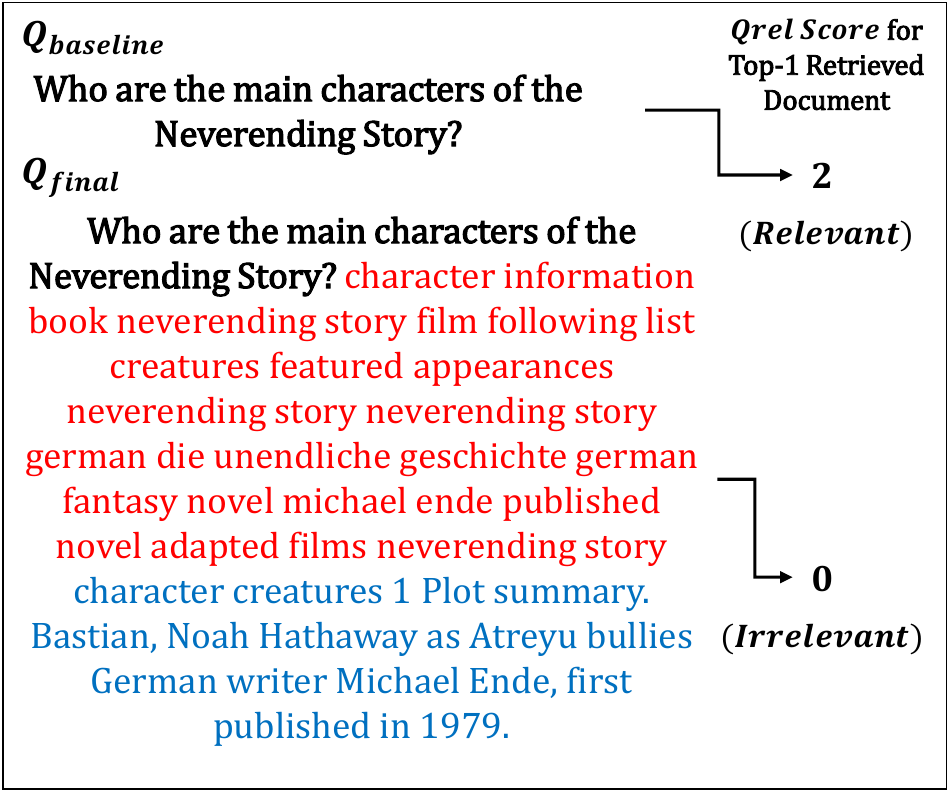}
\caption{Failure case of \textit{GuideCQR} for reformulating conversational query, where the system generates irrelevant keywords and answers with regard to the $Q_{baseline}$.}
\vspace{-4mm}
\label{fig:failure}
\end{figure}

\paragraph{Failure Case Study} 
We analyze a representative failure case of our proposed method by sampling the final query in the CAsT-19 dataset. 
Following previous works~\cite{mao2023large}, we first set the relevance threshold for CAsT-19 at 1, meaning that we consider retrieved documents with a score of 1 or higher relevant to the query.
Therefore, in the sampling failure case process, if the top-ranked (first position) retrieved document has a relevance score of 0, we classify it as a failure case.
As shown in Figure~\ref{fig:failure}, when retrieving documents using $Q_{final}$, the relevance score drops from 2 to 0 compared to $Q_{baseline}$, indicating a shift from relevant to irrelevant content.
In $Q_{baseline}$, the query pertains to the main character of \textit{The NeverEnding Story}, but the red-marked keywords and blue-marked answers in the retrieved documents refer to actual authors or actors, leading to a failure case.
This observation suggests that performance tends to degrade when unrelated keywords and answers are introduced into the query, deviating from the original context of $Q_{baseline}$.

\paragraph{Adaptability of \textit{GuideCQR}} 

\textit{GuideCQR} is a highly effective framework as it operates as a query expansion method, making it adaptable for use with other methods or queries simultaneously. To demonstrate its effectiveness, we conduct experiments using human-rewritten queries, $RawQuery$ and $Q_{baseline}$
with different generator models from the CAsT-19 dataset. We create $Q_{baseline}$ using GPT-3.5-turbo-16k to derive the final performance of GuideCQR. Furthermore, we conduct experiments with EXAONE-3.5-7.8B-Instruct~\cite{exaone-3.5}, Mistral-7B-Instruct-v0.3~\cite{jiang2023mistral}, and Qwen2.5-7B-Instruct~\cite{qwen2.5}. As shown in Table ~\ref{tab:adaptability}, $RawQuery$, human-rewritten queries and $Q_{baseline}$ with different generators show performance improvements. This demonstrates that our framework can be applied to any query or CQR method to achieve enhanced performance. 

\section{Conclusion}
We present \textit{GuideCQR}, a novel query reformulation framework utilizing initially retrieved documents from query set for conversational search. \textit{GuideCQR} effectively reformulates conversational queries to be more retriever-friendly by extracting meaningful information from these guided documents. 
%\textit{GuideCQR} significantly enhances retrieval results, offering a meaningful advancement in conversational search systems. 
Experimental results across various datasets and metrics confirm the capability of \textit{GuideCQR} over human rewritten query and previous CQR methods. We also show that GuideCQR can be effectively adapted to various types of re-written queries.

\section*{Limitations}
Our proposed framework generally takes more time to reformulate conversational queries compared to conventional CQR methods. In particular, the stages involving keyword augmentation and expected answer generation tend to be more time-consuming in \textit{GuideCQR}, especially as the keyword span length and the number of documents increase. Future research will need to address these inefficiencies to reduce the overall processing time.

Moreover, the optimal keyword span length and number of documents may vary depending on the dataset. While incorporating more signals can enhance search results, the ideal amount of signals across diverse datasets remains undetermined. Thus, optimizing signal usage for different datasets represents a promising avenue for future exploration.

Despite these limitations, we believe our framework offers a novel perspective on addressing CQR challenges. As a result, we expect it will positively influence the future development of CQR methods.

\section*{Ethics Statement}
This research leverages multiple publicly available datasets for conversational query reformulation. We have accurately cited all the papers and sources used in our study. We intend to publish the final query and the code including the pre-trained model for our proposed framework once the paper is accepted.

\section*{Acknowledgement}
This research was supported by Institute for Information \& Communications Technology Planning \& Evaluation (IITP) through the Korea government (MSIT) under Grant No. 2021-0-01341 (Artificial Intelligence Graduate School Program (Chung-Ang University)).

\bibliography{main}

\begin{thebibliography}{36}
\expandafter\ifx\csname natexlab\endcsname\relax\def\natexlab#1{#1}\fi

\bibitem[{Anantha et~al.(2021)Anantha, Vakulenko, Tu, Longpre, Pulman, and Chappidi}]{anantha-etal-2021-open}
Raviteja Anantha, Svitlana Vakulenko, Zhucheng Tu, Shayne Longpre, Stephen Pulman, and Srinivas Chappidi. 2021.
\newblock \href {https://doi.org/10.18653/v1/2021.naacl-main.44} {Open-domain question answering goes conversational via question rewriting}.
\newblock In \emph{Proceedings of the 2021 Conference of the North American Chapter of the Association for Computational Linguistics: Human Language Technologies}, pages 520--534, Online. Association for Computational Linguistics.

\bibitem[{Campos et~al.(2016)Campos, Nguyen, Rosenberg, Song, Gao, Tiwary, Majumder, Deng, and Mitra}]{Campos2016MSMA}
Daniel~Fernando Campos, Tri Nguyen, Mir Rosenberg, Xia Song, Jianfeng Gao, Saurabh Tiwary, Rangan Majumder, Li~Deng, and Bhaskar Mitra. 2016.
\newblock \href {https://api.semanticscholar.org/CorpusID:1289517} {Ms marco: A human generated machine reading comprehension dataset}.
\newblock \emph{ArXiv}, abs/1611.09268.

\bibitem[{Cheng et~al.(2024)Cheng, Mao, and Dou}]{cheng-etal-2024-interpreting}
Yiruo Cheng, Kelong Mao, and Zhicheng Dou. 2024.
\newblock \href {https://aclanthology.org/2024.acl-long.159} {Interpreting conversational dense retrieval by rewriting-enhanced inversion of session embedding}.
\newblock In \emph{Proceedings of the 62nd Annual Meeting of the Association for Computational Linguistics (Volume 1: Long Papers)}, pages 2879--2893, Bangkok, Thailand. Association for Computational Linguistics.

\bibitem[{Dalton et~al.(2021)Dalton, Xiong, and Callan}]{dalton2021cast}
Jeffrey Dalton, Chenyan Xiong, and Jamie Callan. 2021.
\newblock Cast 2020: The conversational assistance track overview.

\bibitem[{Dalton et~al.(2020)Dalton, Xiong, Kumar, and Callan}]{dalton2020cast}
Jeffrey Dalton, Chenyan Xiong, Vaibhav Kumar, and Jamie Callan. 2020.
\newblock Cast-19: A dataset for conversational information seeking.
\newblock In \emph{Proceedings of the 43rd International ACM SIGIR Conference on Research and Development in Information Retrieval}, pages 1985--1988.

\bibitem[{Devlin et~al.(2019)Devlin, Chang, Lee, and Toutanova}]{devlin-etal-2019-bert}
Jacob Devlin, Ming-Wei Chang, Kenton Lee, and Kristina Toutanova. 2019.
\newblock \href {https://doi.org/10.18653/v1/N19-1423} {{BERT}: Pre-training of deep bidirectional transformers for language understanding}.
\newblock In \emph{Proceedings of the 2019 Conference of the North {A}merican Chapter of the Association for Computational Linguistics: Human Language Technologies, Volume 1 (Long and Short Papers)}, pages 4171--4186, Minneapolis, Minnesota. Association for Computational Linguistics.

\bibitem[{Elgohary et~al.(2019)Elgohary, Peskov, and Boyd-Graber}]{elgohary2019can}
Ahmed Elgohary, Denis Peskov, and Jordan Boyd-Graber. 2019.
\newblock Can you unpack that? learning to rewrite questions-in-context.
\newblock \emph{Can You Unpack That? Learning to Rewrite Questions-in-Context}.

\bibitem[{Gao et~al.(2022)Gao, Xiong, Bennett, and Craswell}]{gao2023neural}
Jianfeng Gao, Chenyan Xiong, Paul Bennett, and Nick Craswell. 2022.
\newblock \emph{Neural approaches to conversational information retrieval}.
\newblock Springer.

\bibitem[{Grootendorst(2020)}]{grootendorst2020keybert}
Maarten Grootendorst. 2020.
\newblock \href {https://doi.org/10.5281/zenodo.4461265} {Keybert: Minimal keyword extraction with bert.}

\bibitem[{Jagerman et~al.(2023)Jagerman, Zhuang, Qin, Wang, and Bendersky}]{jagerman2023query}
Rolf Jagerman, Honglei Zhuang, Zhen Qin, Xuanhui Wang, and Michael Bendersky. 2023.
\newblock Query expansion by prompting large language models.
\newblock \emph{arXiv preprint arXiv:2305.03653}.

\bibitem[{Jang et~al.(2024)Jang, Lee, Bae, Lee, and Jung}]{jang-etal-2024-itercqr}
Yunah Jang, Kang-il Lee, Hyunkyung Bae, Hwanhee Lee, and Kyomin Jung. 2024.
\newblock \href {https://doi.org/10.18653/v1/2024.naacl-long.449} {{I}ter{CQR}: Iterative conversational query reformulation with retrieval guidance}.
\newblock In \emph{Proceedings of the 2024 Conference of the North American Chapter of the Association for Computational Linguistics: Human Language Technologies (Volume 1: Long Papers)}, pages 8121--8138, Mexico City, Mexico. Association for Computational Linguistics.

\bibitem[{Jiang et~al.(2023)Jiang, Sablayrolles, Mensch, Bamford, Chaplot, Casas, Bressand, Lengyel, Lample, Saulnier et~al.}]{jiang2023mistral}
Albert~Q Jiang, Alexandre Sablayrolles, Arthur Mensch, Chris Bamford, Devendra~Singh Chaplot, Diego de~las Casas, Florian Bressand, Gianna Lengyel, Guillaume Lample, Lucile Saulnier, et~al. 2023.
\newblock Mistral 7b.
\newblock \emph{arXiv preprint arXiv:2310.06825}.

\bibitem[{Johnson et~al.(2019)Johnson, Douze, and J{\'e}gou}]{johnson2019billion}
Jeff Johnson, Matthijs Douze, and Herv{\'e} J{\'e}gou. 2019.
\newblock Billion-scale similarity search with gpus.
\newblock \emph{IEEE Transactions on Big Data}, 7(3):535--547.

\bibitem[{Kim and Kim(2022)}]{kim2022saving}
Sungdong Kim and Gangwoo Kim. 2022.
\newblock Saving dense retriever from shortcut dependency in conversational search.
\newblock In \emph{Proceedings of the 2022 Conference on Empirical Methods in Natural Language Processing}, pages 10278--10287.

\bibitem[{Lin et~al.(2021)Lin, Ma, Lin, Yang, Pradeep, and Nogueira}]{lin2021pyserini}
Jimmy Lin, Xueguang Ma, Sheng-Chieh Lin, Jheng-Hong Yang, Ronak Pradeep, and Rodrigo~Frassetto Nogueira. 2021.
\newblock Pyserini: A python toolkit for reproducible information retrieval research with sparse and dense representations.
\newblock In \emph{SIGIR}.

\bibitem[{Lin et~al.(2021b)Lin, Yang, and Lin}]{sheng2021bin}
Sheng-Chieh Lin, Jheng-Hong Yang, and Jimmy Lin. 2021b.
\newblock Contextualized query embeddings for conversational search.
\newblock In \emph{In Proceedings of the 2021 Conference on Empirical Methods in Natural Language Processing}, pages 1004--1015.

\bibitem[{Lin et~al.(2020)Lin, Yang, Nogueira, Tsai, Wang, and Lin.}]{sheng2020conversational}
Sheng-Chieh Lin, Jheng-Hong Yang, Rodrigo Nogueira, Ming-Feng Tsai, Chuan-Ju Wang, and Jimmy Lin. 2020.
\newblock Conversational question reformulation via sequence-to-sequence architectures and pretrained language models.

\bibitem[{Mao et~al.(2023)Mao, Dou, Mo, Hou, Chen, and Qian}]{mao2023large}
Kelong Mao, Zhicheng Dou, Fengran Mo, Jiewen Hou, Haonan Chen, and Hongjin Qian. 2023.
\newblock Large language models know your contextual search intent: A prompting framework for conversational search.
\newblock In \emph{Findings of the Association for Computational Linguistics: EMNLP 2023}, pages 1211--1225.

\bibitem[{Mo et~al.(2024)Mo, Ghaddar, Mao, Rezagholizadeh, Chen, Liu, and Nie}]{mo2024chiq}
Fengran Mo, Abbas Ghaddar, Kelong Mao, Mehdi Rezagholizadeh, Boxing Chen, Qun Liu, and Jian-Yun Nie. 2024.
\newblock Chiq: Contextual history enhancement for improving query rewriting in conversational search.
\newblock \emph{arXiv preprint arXiv:2406.05013}.

\bibitem[{Mo et~al.(2023{\natexlab{a}})Mo, Mao, Zhu, Wu, Huang, and Nie}]{mo2023convgqr}
Fengran Mo, Kelong Mao, Yutao Zhu, Yihong Wu, Kaiyu Huang, and Jian-Yun Nie. 2023{\natexlab{a}}.
\newblock Convgqr: Generative query reformulation for conversational search.
\newblock In \emph{Proceedings of the 61st Annual Meeting of the Association for Computational Linguistics (Volume 1: Long Papers)}, pages 4998--5012.

\bibitem[{Mo et~al.(2023{\natexlab{b}})Mo, Nie, Huang, Mao, Zhu, Li, and Liu}]{mo2023learning}
Fengran Mo, Jian-Yun Nie, Kaiyu Huang, Kelong Mao, Yutao Zhu, Peng Li, and Yang Liu. 2023{\natexlab{b}}.
\newblock Learning to relate to previous turns in conversational search.
\newblock In \emph{Proceedings of the 29th ACM SIGKDD Conference on Knowledge Discovery and Data Mining}, pages 1722--1732.

\bibitem[{Qu et~al.(2020)Qu, Yang, Chen, Qiu, Croft, and Iyyer}]{qu2020open}
Chen Qu, Liu Yang, Cen Chen, Minghui Qiu, W~Bruce Croft, and Mohit Iyyer. 2020.
\newblock Open-retrieval conversational question answering.
\newblock In \emph{Proceedings of the 43rd International ACM SIGIR conference on research and development in Information Retrieval}, pages 539--548.

\bibitem[{Reimers and Gurevych(2019)}]{reimers2019sentence}
Nils Reimers and Iryna Gurevych. 2019.
\newblock Sentence-bert: Sentence embeddings using siamese bert-networks.
\newblock In \emph{Proceedings of the 2019 Conference on Empirical Methods in Natural Language Processing and the 9th International Joint Conference on Natural Language Processing (EMNLP-IJCNLP)}, pages 3982--3992.

\bibitem[{Rennie et~al.(2017)Rennie, Marcheret, Mroueh, Ross, and Goel}]{rennie2017self}
Steven~J Rennie, Etienne Marcheret, Youssef Mroueh, Jerret Ross, and Vaibhava Goel. 2017.
\newblock Self-critical sequence training for image captioning.
\newblock In \emph{Proceedings of the IEEE conference on computer vision and pattern recognition}, pages 7008--7024.

\bibitem[{Research(2024)}]{exaone-3.5}
LG~AI Research. 2024.
\newblock Exaone 3.5: Series of large language models for real-world use cases.
\newblock \emph{arXiv preprint arXiv:https://arxiv.org/abs/2412.04862}.

\bibitem[{Robertson et~al.(2009)Robertson, Zaragoza et~al.}]{robertson2009probabilistic}
Stephen Robertson, Hugo Zaragoza, et~al. 2009.
\newblock The probabilistic relevance framework: Bm25 and beyond.
\newblock \emph{Foundations and Trends{\textregistered} in Information Retrieval}, 3(4):333--389.

\bibitem[{Team(2024)}]{qwen2.5}
Qwen Team. 2024.
\newblock \href {https://qwenlm.github.io/blog/qwen2.5/} {Qwen2.5: A party of foundation models}.

\bibitem[{Vakulenko et~al.(2021a)Vakulenko, Longpre, Tu, and Anantha}]{svitlana2021ain}
Svitlana Vakulenko, Shayne Longpre, Zhucheng Tu, and Raviteja Anantha. 2021a.
\newblock Question rewriting for conversational question answering.
\newblock In \emph{In Proceedings of the 14th ACM International Conference on Web Search and Data Mining}, pages 355--363.

\bibitem[{Van~Gysel and de~Rijke(2018)}]{van2018pytrec_eval}
Christophe Van~Gysel and Maarten de~Rijke. 2018.
\newblock Pytrec\_eval: An extremely fast python interface to trec\_eval.
\newblock In \emph{The 41st International ACM SIGIR Conference on Research \& Development in Information Retrieval}, pages 873--876.

\bibitem[{Voskarides et~al.(2020{\natexlab{a}})Voskarides, Li, Ren, Kanoulas, and de~Rijke}]{voskarides2020query}
Nikos Voskarides, Dan Li, Pengjie Ren, Evangelos Kanoulas, and Maarten de~Rijke. 2020{\natexlab{a}}.
\newblock Query resolution for conversational search with limited supervision.
\newblock In \emph{Proceedings of the 43rd International ACM SIGIR conference on research and development in Information Retrieval}, pages 921--930.

\bibitem[{Voskarides et~al.(2020{\natexlab{b}})Voskarides, Li, Ren, Kanoulas, and de~Rijke}]{nikos2020in}
Nikos Voskarides, Dan Li, Pengjie Ren, Evangelos Kanoulas, and Maarten de~Rijke. 2020{\natexlab{b}}.
\newblock Query resolution for conversational search with limited supervision.
\newblock In \emph{In Proceedings of the 43rd International ACM SIGIR conference on research and development in Information Retrieval}, pages 921--930.

\bibitem[{Wang et~al.(2023)Wang, Yang, and Wei}]{wang2023query2doc}
Liang Wang, Nan Yang, and Furu Wei. 2023.
\newblock Query2doc: Query expansion with large language models.
\newblock In \emph{Proceedings of the 2023 Conference on Empirical Methods in Natural Language Processing}, pages 9414--9423.

\bibitem[{Wu et~al.(2022)Wu, Luan, Rashkin, Reitter, Hajishirzi, Ostendorf, and Tomar}]{wu2022conqrr}
Zeqiu Wu, Yi~Luan, Hannah Rashkin, David Reitter, Hannaneh Hajishirzi, Mari Ostendorf, and Gaurav~Singh Tomar. 2022.
\newblock Conqrr: Conversational query rewriting for retrieval with reinforcement learning.
\newblock In \emph{Proceedings of the 2022 Conference on Empirical Methods in Natural Language Processing}, pages 10000--10014.

\bibitem[{Xiong et~al.(2021)Xiong, Xiong, Li, Tang, Liu, Bennett, Ahmed, and Overwijk}]{xiong2020approximate}
Lee Xiong, Chenyan Xiong, Ye~Li, Kwok{-}Fung Tang, Jialin Liu, Paul~N. Bennett, Junaid Ahmed, and Arnold Overwijk. 2021.
\newblock \href {https://openreview.net/forum?id=zeFrfgyZln} {Approximate nearest neighbor negative contrastive learning for dense text retrieval}.
\newblock In \emph{9th International Conference on Learning Representations, {ICLR} 2021, Virtual Event, Austria, May 3-7, 2021}. OpenReview.net.

\bibitem[{Ye et~al.(2023)Ye, Fang, Li, and Yilmaz}]{ye-etal-2023-enhancing}
Fanghua Ye, Meng Fang, Shenghui Li, and Emine Yilmaz. 2023.
\newblock \href {https://doi.org/10.18653/v1/2023.findings-emnlp.398} {Enhancing conversational search: Large language model-aided informative query rewriting}.
\newblock In \emph{Findings of the Association for Computational Linguistics: EMNLP 2023}, pages 5985--6006, Singapore. Association for Computational Linguistics.

\bibitem[{Yu et~al.(2020)Yu, Liu, Yang, Xiong, Bennett, Gao, and Liu}]{yu2020few}
Shi Yu, Jiahua Liu, Jingqin Yang, Chenyan Xiong, Paul Bennett, Jianfeng Gao, and Zhiyuan Liu. 2020.
\newblock Few-shot generative conversational query rewriting.
\newblock In \emph{Proceedings of the 43rd International ACM SIGIR conference on research and development in Information Retrieval}, pages 1933--1936.

\end{thebibliography}
\clearpage
\appendix

\section{Implementation Details}
\label{app:details}
To re-rank the guided documents as in section~\ref{sec:rerank}, we select the \textit{ember-v1}~\footnote{https://huggingface.co/llmrails/ember-v1} and \textit{mxbai-embed-large-v1}~\footnote{https://huggingface.co/mixedbread-ai/mxbai-embed-large-v1} models as the first and second re-ranking Sentence-Transformer models, respectively. And we use OpenAI \textit{text-embedding-3-small} for generating embeddings for keywords and answers in the filtering stage in \textit{GuideCQR}. We use different embedding models at each stage to mitigate potential biases caused by a single model. We use KeyBERT~\citep{grootendorst2020keybert} for augmenting keywords. And we use \textit{RoBERTa-Base-Squad2} ~\footnote{https://huggingface.co/deepset/roberta-base-squad2} for generating the expected answer. We use the most common query among the five candidate queries provided by the baseline ~\cite{mao2023large} during the re-ranking, answer generation and filtering process in CAsT datasets. 
This is because ~\cite{mao2023large} provides a set of five candidate queries when generating their queries, so we make representative among them. And in augmenting keyword process, the number of keywords extracted can be fewer than the span length because the number of keywords reasonable in each document can be less than the length of the keyword span. This means that the keyword span length is maximum number of keywords for each document. We use \textit{roberta-large-mnli} ~\footnote{https://huggingface.co/FacebookAI/roberta-large-mnli} for measuring entailment score between augmented keywords and $Q_{baseline}$. In the query relevance analysis, we ensure a fair comparison by standardizing the number of documents in each set to 1,700 before conducting the experiment. We leverage Faiss index~\cite{johnson2019billion} for ANCE retriever.
We conduct our experiments using an AMD EPYC 7313 CPU (3.0 GHz) paired with four NVIDIA RTX 4090 GPUs. We use Python 3.11.5 and PyTorch 2.3.1 for the software environment.

\paragraph{Selecting FilterScore} 
We experiment with various $FilterScore$ settings to find the optimal hyperparameter with the validation set for the proposed method. The number of documents for keywords and expected answers and optimal keyword span length varies by dataset; these values were chosen to improve retrieval results and enhance query quality for the retriever. Therefore, before filtering, we use top-4 span 15 keywords and top-10 answers for CAsT-19, and top-5 span 5 keywords and top-10 answers for CAsT-20 and top-1 span 10 keywords and top-10 answers for QReCC. And \textit{GuideCQR} uses these keywords and answers as our final query set. We select the $FilterScore$ for each keyword and answer by experimenting with validation sets for the datasets to find the optimal scores. We achieve our final results using a $FilterScore$ of (1, 1.9) for each keyword and answer in CAsT-19, (0.1, 1.95) in CAsT-20, and (0.5, 9) in QReCC.

\begin{table*}[h]
    \centering
    \small
    \begin{tabular}{l|ccc|ccc|ccc|c}
        \toprule
        & \multicolumn{3}{c|}{CAsT-19} & \multicolumn{3}{c|}{CAsT-20} & \multicolumn{3}{c|}{QReCC} & \textbf{Avg} \\
        \cmidrule(lr){2-4} \cmidrule(lr){5-7} \cmidrule(lr){8-10} 
        \textbf{Methods} & MRR & NDCG@3 & R@10 & MRR & NDCG@3 & R@10 & MRR & NDCG@3 & R@10 & \\
        \midrule
        RawQuery & 31.9 & 13.1 & 3.8 & 9.9 & 6.8 & 2.5 & 6.5 & 5.5 & 11.1 & 10.1 \\
        $Q_{baseline}$ & 59.9 & 29.5 & 10.3 & 30.5 & 19.0 & 10.7 & 46.9 & 55.0 & 65.5 & 36.4 \\
        LLM4CS & \underline{67.5} & \underline{41.9} & \underline{11.5} & \textbf{53.7} & \textbf{38.8} & \textbf{20.5} & \underline{47.8} & \underline{45.0} & \underline{69.1} & \underline{44.0} \\
        \textbf{GuideCQR (ours)} & \textbf{75.4} & \textbf{49.7} & \textbf{14.0} & \underline{49.1} & \underline{34.4} & \underline{16.4} & \textbf{51.2} & \textbf{58.6} & \textbf{69.0} & \textbf{46.4} \\
        \midrule
        Human Rewrite & 61.3 & 30.9 & 10.9 & 37.7 & 23.9 & 14.2 & 39.7 & 36.2 & 62.5 & 35.3 \\
        \bottomrule
    \end{tabular}
    \caption{Performance comparison on CAsT-19, CAsT-20, and QReCC datasets using sparse retriever (BM25). We present MRR, NDCG@3, R@10, and the average of all scores for each method. The best results are in bold, and the second best are underlined.}
    \label{tab:sparse}
    \vspace{-3mm}
\end{table*}
\section{Result using Sparse Retrieval}
We provide the performance of \textit{GuideCQR} using sparse retrieval. This result is not included in the main table because prior studies on the CAsT series have predominantly focused on dense retrieval. Consequently, retrieval results for baseline methods are unavailable and the rewritten queries are not provided either. Thus, we conduct experiments with the available resources. We perform retrieval using BM25~\cite{robertson2009probabilistic}, and indexing is performed via Pyserini~\cite{lin2021pyserini}. For sparse retrieval in LLM4CS~\cite{mao2023large}, we concatenate the rewritten query and response of generated LLM4CS final query. 

As shown in Table~\ref{tab:sparse}, \textit{GuideCQR} significantly enhances performance metrics compared to the $Q_{baseline}$ across all datasets. \textit{GuideCQR} achieves state-of-the-art performance in terms of the average score. Similar to earlier observations, the scores for CAsT-20 are lower due to the unique characteristics of the CAsT datasets.

\section{Dataset Statistics}
\paragraph{CAsT} Table ~\ref{tab:cast} presents statistics for the CAsT datasets, including the number of conversations, queries, and documents. 

\begin{table}[h]
\centering
\begin{tabular}{lcccc}
\hline
 & \textbf{CAsT-19} & \textbf{CAsT-20} \\
\hline
\# Conv      & 50  & 25  \\
\# Questions & 479 & 216 \\
\# Documents & 38M & 38M \\
\hline
\end{tabular}
\caption{Dataset Statistics for CAsT-19 and CAsT-20.}
\label{tab:cast}
\end{table}

\paragraph{QReCC} 
Table~\ref{tab:qrecc} shows the statistics for the QReCC dataset, which contains approximately 54M documents. We use a final test set of 8,209 questions, following ~\cite{ye-etal-2023-enhancing}, after removing invalid gold passage labels. This test set includes 6,396 questions for QuAC-Conv, 1,442 for NQ-Conv, and 371 for TREC-Conv.

\begin{table}[h!]
\centering
\begin{tabular}{lccc}
\hline
 & \textbf{Train} & \textbf{Dev} & \textbf{Test} \\
\hline
\textbf{QReCC} & & & \\
\# Conv & 8,823 & 2,000 & 2,775 \\
\# Questions & 51,928 & 11,573 & 16,451 \\
\hline
\textbf{QuAC-Conv} & & & \\
\# Conv & 6,008 & 1,300 & 1,816 \\
\# Questions & 41,395 & 8,965 & 12,389 \\
\hline
\textbf{NQ-Conv} & & & \\
\# Conv & 2,815 & 700 & 879 \\
\# Questions & 10,533 & 2,608 & 3,314 \\
\hline
\textbf{TREC-Conv} & & & \\
\# Conv & 0 & 0 & 80 \\
\# Questions & 0 & 0 & 748 \\
\hline
\end{tabular}
\caption{Dataset Statistics for QReCC, QuAC-Conv, NQ-Conv, and TREC-Conv.}
\label{tab:qrecc}
\end{table}

\begin{comment}
    \section{Performance among Amount of Initial Guided documents}

\input{tables/amount_of_documents}

We evaluate the impact and robustness of each step in the \textit{GuideCQR} setup. 
Initially, we adjust the number of guided documents to observe the proper quantity and present the results in Table~\ref{tab:7}. 
Our findings demonstrate that increasing the number of guided documents consistently enhances performance. 
However, retrieving an excessive number of documents leads to longer inference times.
Therefore, we limit the number of initial documents to 2,000 for our main experiment.
%Hence, considering the computation cost, we decided to use 2000 documents, which is the point on the dev set where there is no additional performance improvement in at least one metric. We also used the same criteria for keyword span length.  이 부분 추가 
\end{comment}

\section{Performance among Keyword Span Length}
We also investigate the impact of varying keyword spans in augmenting the keyword stage and present the result in Table ~\ref{tab:2}. We observe that increasing the keyword span generally enhances performance to a certain extent. However, we find that too many signals can degrade performance due to inclusion of the redundant information in the query.
\begin{table}[h]
    \centering
    \resizebox{\columnwidth}{!}{ 
    \begin{tabular}{l|cc|cc}
        \toprule
        & \multicolumn{2}{c|}{CAsT-19} & \multicolumn{2}{c}{CAsT-20} \\
        \cmidrule(lr){2-3} \cmidrule(lr){4-5}
         & MRR & NDCG@3 & MRR & NDCG@3 \\
        \midrule
        $Q_{baseline}$ & 72.2 & 45.2 & 53.4 & 37.2 \\
        5 span & 76.9 & 49.9 & 57.7 & 41.5 \\
        10 span & 78.8 & 52.0 & 53.7 & 39.8 \\
        15 span & 81.0 & 52.2 & 54.1 & 40.0 \\
        20 span & 80.1 & 52.0 & 54.6 & 39.8 \\
        \bottomrule
    \end{tabular}
    }
    \caption{Performance among the different numbers of span for extracting keywords. We augment all keywords from top-4 documents.}
    \label{tab:2}
    \vspace{-3mm}
\end{table}

\section{Performance for removing duplicated keywords}

We conduct experiments by removing duplicate keywords, as non-duplicated keywords may lead to improved retrieval performance. Table~\ref{tab:unique} shows that duplicated keywords performed slightly better than unique ones, so we keep them duplicated. 

\begin{table}[h]
    \centering
    \begin{tabular}{l|cc}
        \toprule
         & MRR & NDCG@3 \\
        \midrule
        Original & 81.0 & 52.2 \\
        Duplicate\_removed & 78.3 & 49.9 \\
        \bottomrule
    \end{tabular}
    \caption{Evaluation on the CAsT-19 dataset with duplicate-removed keywords. We augment span 15 keywords from top-4 documents.}
    \label{tab:unique}
\end{table}

\section{Process of generating $Q_{baseline}$ of CAsT}
We generate $Q_{baseline}$ of CAsT datasets by utilizing the LLM4CS REW GitHub code which employs GPT-3.5-turbo-16k for prompting ~\cite{mao2023large}. As illustrated in Table ~\ref{tab:8}, the prompt consists of three parts: instruction, demonstration, tail instruction. The final prompt is a concatenation of these three components. Using this prompt, \textit{GuideCQR} generate $Q_{baseline}$ by refining $RawQuery$ with $History$, resolving any coreferences or omissions. 

\section{Dialog history usage in \textit{GuideCQR}}
In conversational search, utilizing dialog history is crucial. \textit{GuideCQR} leverages the conversation history in two key ways. First, we use dialog history when generating $Q_{baseline}$, as it helps refine the current query by resolving any ambiguities, coreferences, or omissions from previous interactions. Additionally, we incorporate history in filtering augmented keywords and expected answers. By measuring the similarity with the history, we can select the most relevant keywords or answers. Through these methods, \textit{GuideCQR} effectively integrates history, demonstrating its capability as a conversational search system.

\section{Query Sample}

We present example queries reformulated by our method and by humans in Table ~\ref{tab:cast_queries}. While our final query, $Q_{final}$, may not appear intuitive from a human perspective, it yields better search results during the retrieval stage.

\section{Samples of Positive Queries on Datasets}
We provide positive query sample of each dataset. For CAsT, if the top-ranked (first position) retrieved document has a relevance score of 1 or higher, we classify it as a positive case. For the QReCC dataset, we refer to a retrieval result as a positive case if the gold passage is included in the top-1 result. Table~\ref{tab:cast19}, Table~\ref{tab:cast20} and Table~\ref{tab:qrecc_sample} shows positive query sample of CAsT-19, CAsT-20, and QReCC dataset.

\begin{table*}[h]
\centering
\begin{tabular}{|l|p{0.77\textwidth}|}
\hline
\textbf{RawQuery} & What are its other competitors? \\
\hline
\textbf{Human-rewritten} & What are Netflix's other competitors than Blockbuster? \\
\hline
\textbf{$Q_{baseline}$} & What are Netflix's other competitors? \\
\hline
\textbf{$Q_{final}$} & What are Netflix's other competitors? netflix competitors subscription streaming amazon instant hulu plus amazon netflix competes viaplay hbo nordic cmore asia netflix competes hooq sky demand competitors apple itunes amazon cinemanow hulu lovefilm google netflix fandangonow ultraflix voddler competitors Amazon Instant Video Service and the Hulu Plus service Presto, Stan and Quickflix FandangoNow, UltraFlix, and Voddler : Competitors brick and mortar video rental stores \\
\hline
\end{tabular}
\caption{Query sample for CAsT-19 (Conv\_id: 49, Turn\_id: 5)}
\label{tab:cast_queries}
\end{table*}

\begin{table*}[h]
\centering
\begin{tabular}{|l|p{0.79\textwidth}|}
\hline
\textbf{Instruction} & For an information-seeking dialog, please help reformulate the question into a rewrite that can fully express the user's information needs without the need for context. \\ 
\hline
\textbf{Demonstration} & I will give you several example multi-turn dialogs, where each turn contains a question, a response, and a rewrite that you need to generate. \\ 
\hline
\textbf{Tail Instruction} & Now, you should give me the rewrite of the **Current Question** under the **Context**. The output format should always be: Rewrite: \$Rewrite. Note that you should always try to rewrite it. Never ask for clarification or say you don't understand it in the generated rewrite. Go ahead! \\ 
\hline
\end{tabular}
\caption{The prompt for generating $Q_{baseline}$ is composed of three components: instruction, demonstration, and tail instruction.}
\label{tab:8}
\end{table*}

\clearpage

\begin{table*}[h]
\centering
\renewcommand{\arraystretch}{1.5}
\setlength{\tabcolsep}{8pt}

\begin{tabular}{|p{3cm}|p{12cm}|}
\hline
\multicolumn{2}{|c|}{\textbf{\large CAsT-19 Dataset}} \\ \hline
\textbf{Previous Turns} & 
Query: What is throat cancer? \newline
Response: “” \newline
Query: Is it treatable? \newline
Response: “” \newline
Query: Tell me about lung cancer. \newline
Response: “” \\ \hline
\textbf{{Original Query}} & What are its symptoms? \\ \hline
\textbf{Human Rewrite} & What are lung cancer's symptoms? \\ \hline
\textbf{LLM} & What are the symptoms of lung cancer? \\ \hline
\textbf{GuideCQR Query} & What are the \textcolor{red}{symptoms of lung cancer}? \textcolor{red}{signs symptoms lung cancer symptoms lung cancer signs symptoms lung cancer cough signs symptoms lung cancer cough} (hemoptysis), \textcolor{red}{breath}, wheezing, hoarseness coughing; pain \textcolor{red}{chest}, \textcolor{red}{coughing} \\ \hline
\textbf{Gold Passage (4)} & \textcolor{red}{Signs} and \textcolor{red}{symptoms} of \textcolor{red}{lung cancer} typically occur only when the disease is advanced. \textcolor{red}{Signs and symptoms} of \textcolor{red}{lung cancer} may include: \textcolor{red}{A new cough} that doesn't go away. Changes in a chronic \textcolor{red}{cough} or smoker's \textcolor{red}{cough}. \textcolor{red}{Coughing} up blood, even a small amount. Shortness of \textcolor{red}{breath}. \textcolor{red}{Chest} pain. \\ \hline
\end{tabular}

\caption{Reformulated Queries by GuideCQR on CAsT-19 Dataset. Overlapping words between \textbf{GuideCQR Query} and \textbf{Gold Passage (4)} are marked in \textcolor{red}{red}. \textbf{Gold Passage(4)} notates that the document has query relevance score of 4 and \textbf{LLM} notates the $Q_{baseline}$.}
\label{tab:cast19}
\end{table*}

\clearpage

\begin{table*}[h!]
\centering
\renewcommand{\arraystretch}{1.5}
\setlength{\tabcolsep}{8pt}

\begin{tabular}{|p{3cm}|p{12cm}|}
\hline
\multicolumn{2}{|c|}{\textbf{\large CAsT-20 Dataset}} \\ \hline
\textbf{Previous Turns} & 
Query: How do you know when your garage door opener is going bad? \newline
Response: Light Socket or Logic Board. If the light is still not coming on, it could be the light socket or the logic board. To determine which one it is, follow the steps below. Unplug the garage door opener, then immediately plug the opener back into the electrical outlet. Listen for the light relay on the logic board to click. If you hear a click, replace the light socket. If you do not hear a click, replace the logic board. There are four causes for the lights on a garage door opener not to come on: 1. Light bulb. 2. Contacts in the light socket. 3. Light socket. \newline
Query: Now it stopped working. Why? \newline
Response: Garage Door Opener Problems. So, when the garage door opener decides to take a day off, it can leave you stuck outside, probably during a rain or snowstorm. Though they may seem complicated, there really are several things most homeowners can do to diagnose and repair opener failures. And, if you are careful not to damage the door or the seal on the bottom of the door, use a flat shovel or similar tool to chip away at the ice. Once you get the door open, clear any water, ice or snow from the spot on the garage floor where the door rests when closed. \\ \hline
\textbf{Original Query} & How much does it cost for someone to fix it? \\ \hline
\textbf{Human Rewrite} & How much does it cost for someone to repair a garage door opener? \\ \hline
\textbf{LLM} & How much does it typically cost to have a professional fix a garage door opener issue? \\ \hline
\textbf{GuideCQR Query} & How much does it typically \textcolor{red}{cost} to have a professional fix a \textcolor{red}{garage door opener} issue? \textcolor{red}{garage door opener repair cost} \textcolor{red}{garage door opener repair cost cost repair} \textcolor{red}{broken} \textcolor{red}{garage door opener} \textcolor{red}{garage door opener repair cost} \textcolor{red}{garage door opener repair cost} fix \textcolor{red}{garage door opener} \\ \hline
\textbf{Gold Passage (4)} & On average, homeowners report paying \$210 to have a \textcolor{red}{garage door opener repair}ed by a handyman. Most homeowners pay between \$170 and \$250 for the \textcolor{red}{cost} of such a \textcolor{red}{repair project}. The minimum reported \textcolor{red}{cost} to \textcolor{red}{repair} a \textcolor{red}{broken} \textcolor{red}{garage door opener} is \$50, while the most \textcolor{red}{cost}ly \textcolor{red}{repair} may amount to \$350. If an electrician is needed to repair faulty wiring, the \textcolor{red}{cost} of the \textcolor{red}{repair} will increase. The minimum reported \textcolor{red}{cost} to \textcolor{red}{repair} a \textcolor{red}{broken} \textcolor{red}{garage door opener} is \$50, while the most costly \textcolor{red}{repair} may amount to \$350. If an electrician is needed to \textcolor{red}{repair} faulty wiring, the \textcolor{red}{cost} of the \textcolor{red}{repair} will increase. \\ \hline
\end{tabular}

\caption{Reformulated Queries by GuideCQR on CAsT-20 Dataset. Overlapping words between \textbf{GuideCQR Query} and \textbf{Gold Passage (4)} are marked in \textcolor{red}{red}. \textbf{Gold Passage(4)} notates that the document has query relevance score of 4 and \textbf{LLM} notates the $Q_{baseline}$.}
\label{tab:cast20}
\end{table*}

\clearpage

\begin{table*}[h]
\centering
\renewcommand{\arraystretch}{1.5}
\setlength{\tabcolsep}{8pt}

\begin{tabular}{|p{3cm}|p{12cm}|}
\hline
\multicolumn{2}{|c|}{\textbf{\large QReCC Dataset}} \\ \hline
\textbf{Previous Turns} & Query: What are the main breeds of goat?, \newline
Answer: Abaza...Zhongwei \\ \hline
\textbf{Original Query} & Tell me about boer goats. \\ \hline
\textbf{Human Rewrite} & Tell me about boer goats. \\ \hline
\textbf{LLM} & What can you tell me about boer goats in relation to the main \textcolor{red}{breeds} of goats? \\ \hline
\textbf{GuideCQR Query} & What can you tell me about \textcolor{red}{boer goats} in relation to the main breeds of \textcolor{red}{goats}? \textcolor{red}{goat boer breed south developed} \\ \hline
\textbf{Gold Answer} & The \textcolor{red}{Boer goat} is a \textcolor{red}{breed} of \textcolor{red}{goat} that was \textcolor{red}{developed} in \textcolor{red}{South} Africa in the early 1900s for meat production. Their name is derived from the Afrikaans (Dutch) word boer, meaning farmer. \\ \hline
\textbf{Gold Passage} & \textcolor{red}{Boer goat} - Wikipedia CentralNotice \textcolor{red}{Boer goat} From Wikipedia, the free encyclopedia Jump to navigation Jump to search A commercial \textcolor{red}{Boer goat} buck Commercial \textcolor{red}{Boer goat} buck \textcolor{red}{Boer goat}, Doe The \textcolor{red}{Boer goat} is a \textcolor{red}{breed} of \textcolor{red}{goat} that was \textcolor{red}{developed} in \textcolor{red}{South} Africa in the early 1900s and is a popular \textcolor{red}{breed} for meat production. Their name is derived from the Afrikaans ( Dutch ) word boer , meaning farmer. Contents 1 Origins and characteristics 2 Doe 3 Crossbreeding 4 Show \textcolor{red}{goats} 5 References 6 External links Origins and characteristics [ edit ] The \textcolor{red}{Boer goat} was probably bred from the indigenous \textcolor{red}{South} African \textcolor{red}{goats} kept by the Namaqua , San , and Fooku tribes, with some crossing of Indian and European bloodlines being possible. They were selected for meat rather than milk production; due to selective breeding and improvement, the \textcolor{red}{Boer goat} has a fast growth rate and excellent carcass qualities, making it one of the most popular \textcolor{red}{breeds} of meat \textcolor{red}{goat} in the world. \textcolor{red}{Boer goats} have a high resistance to disease and adapt well to hot, dry semideserts. United States production is centered in west-central Texas , particularly in and around San Angelo and Menard . The original US breeding stock came from herds located in New Zealand. Only later were they imported directly from Africa \textcolor{red}{Boer goats} commonly have white bodies and distinctive brown heads. Some \textcolor{red}{Boer goats} can be completely brown or white or paint, which means large spots of a different color are on their bodies. Like the Nubian \textcolor{red}{goat} , they possess long, pendulous ears. They are noted for being docile, fast-growing, and having high fertility rates. Does are reported to have superior mothering skills as compared to other \textcolor{red}{breeds}. \\ \hline
\end{tabular}

\caption{Reformulated Queries by GuideCQR on QReCC Dataset. Overlapping words between \textbf{GuideCQR Query} and \textbf{Gold Passage}, \textbf{Gold Answer} are marked in \textcolor{red}{red}. \textbf{LLM} notates the $Q_{baseline}$.}
\label{tab:qrecc_sample}
\end{table*}

\end{document}